\documentclass[letterpaper, 10 pt, conference]{ieeeconf}  % Comment this line out if you need a4paper

\IEEEoverridecommandlockouts                              % This command is only needed if 
\usepackage[bottom=1.95cm, top=1.95cm, left=1.91cm, right=1.91cm]{geometry}
\usepackage[labelformat=simple,labelsep=colon]{caption}
\usepackage{array}
\usepackage{graphics} % for pdf, bitmapped graphics files
\usepackage{amsmath} % assumes amsmath package installed
\usepackage{amssymb}  % assumes amsmath package installed
\usepackage{graphicx}
\usepackage{cite}
\usepackage{booktabs}
\usepackage{multirow}
\usepackage{tabularx}

\newcolumntype{Y}{>{\centering\arraybackslash}X}
\title{\LARGE \bf
Morphology-Independent Facial Expression Imitation

for Human-Face Robots
}

\author{Xu Chen, Rui Gao, Che Sun*, Zhehang Liu, Yuwei Wu, Shuo Yang, Yunde Jia% <-this % stops a space
}

\begin{document}

\maketitle
\thispagestyle{empty}
\pagestyle{empty}

%%%%%%%%%%%%%%%%%%%%%%%%%%%%%%%%%%%%%%%%%%%%%%%%%%%%%%%%%%%%%%%%%%%%%%%%%%%%%%%%
\begin{abstract}

Accurate facial expression imitation on human-face robots is crucial for achieving natural human–robot interaction.  
Most existing methods have achieved photorealistic expression imitation through mapping 2D facial landmarks to a robot's actuator commands. Their imitation of landmark trajectories is susceptible to interference from facial morphology, which would lead to a performance drop.
% To this end, 
In this paper, we propose a morphology-independent expression imitation method that decouples expressions from facial morphology to eliminate morphological influence and produce more realistic expressions for human-face robots. 
Specifically, we construct an expression decoupling module to learn expression semantics by disentangling the expression representation from the morphology representation in a self-supervised manner. We devise an expression transfer module to map the representations to the robot's actuator commands through a learning objective of perceiving expression errors, producing accurate facial expressions based on the learned expression semantics. 
To support experimental validation, a custom-designed and highly expressive human-face robot, namely Pengrui, is developed to serve as an experimental platform for realistic expression imitation.
Extensive experiments demonstrate that our method enables the human-face robot to  reproduce a wide range of human-like expressions effectively. All code and implementation details of the robot will be released. 
\end{abstract}

%%%%%%%%%%%%%%%%%%%%%%%%%%%%%%%%%%%%%%%%%%%%%%%%%%%%%%%%%%%%%%%%%%%%%%%%%%%%%%%%
\section{INTRODUCTION}

Facial expression imitation for human-face robots seeks to faithfully reproduce human facial expressions to enable natural and expressive human–robot interaction, garnering significant attention across research communities such as human-robot interaction, healthcare, and social robotics\cite{pennisi2016autism,10.1145/3568162.3578625,fogelson2022impact,laban2025coping}. 

Most existing methods rely on establishing a direct mapping from either pre-define patterns of facial expressions~\cite{oh2006design,hashimoto2006development,hashimoto2008dynamic} or sparse 2D facial landmarks~\cite{chen2021smile,hu2024human} to the robot’s actuator commands for facial expression imitation. While these approaches have achieved realistic expression imitation under consistent facial morphology, their imitation performance is interfered with by morphological variations. 
This interference  stems from the inherent coupling between the used expression representations (such as 2D facial landmarks) and the underlying facial morphology. 
Existing imitation methods rely on these coupled representations, and they could risk misinterpreting the morphology differences between individuals as expression movements, leading to erroneous actuator commands and distorted robotic expressions. In this paper, we propose to
decouple expressions from facial morphology to eliminate the interference, producing more faithful expressions on human-face robots.

Decoupling expressions from facial morphology is non-trivial for facial expression imitation on human-face robots. 
In biology, morphology-independent facial models remain largely unexplored, and the commonly used electromyography-based analysis techniques~\cite{liang2025largemodelempoweredembodied,hess2013emotional,iacoboni2005neural} still exhibit entanglement between expression signals and individual morphology. Although deep representation learning has demonstrated remarkable capacity for implicit semantic disentanglement in computer vision~\cite{wang2024disentangled,carbonneau2022measuring,hsieh2018learning} and natural language processing~\cite{xu2022compositional,xie2024graph,abbasi2024deciphering},  its application to human-face robotics is impeded by the absence of annotated datasets containing the same expression performed across diverse facial morphology.  Such supervised data  is particularly difficult to acquire because determining whether different expressions are the same is inherently subjective and lacks deterministic ground truth. 

\begin{figure}[t]
    \centering
    \includegraphics[width=\linewidth]{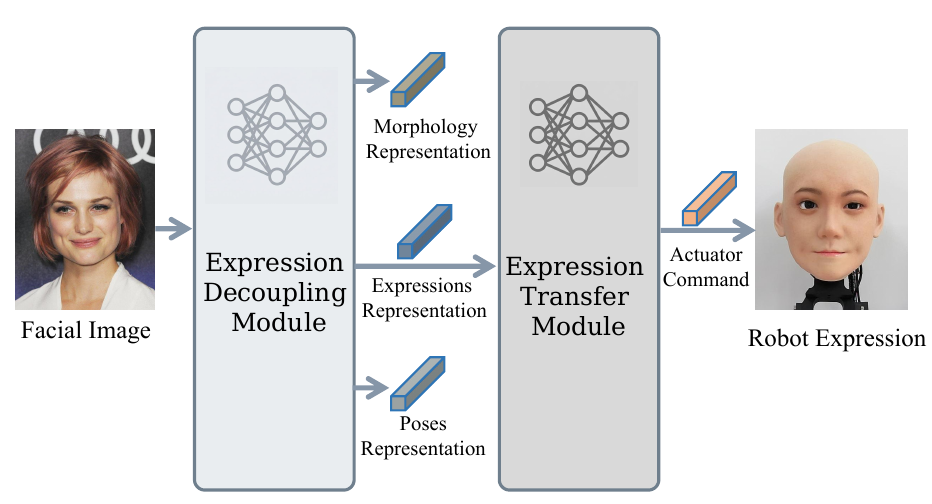} 
    \caption{\textbf{The overview of the morphology-independent facial expression imitation method.} An  facial image is  processed by the expression decoupling module to extract the disentanglemented representations of facial expressions, morphology and poses.  The expression representations are then mapped to actuator commands  via the expression transfer module, actuating Pengrui, i.e., our custom-designed and highl expressive human-face robot, to reproduce the target expression.}
    \label{pipeline_all}
\end{figure}

To address the challenge, we propose a morphology-independent expression imitation method that decouples facial expressions from morphology in a self-supervised manner. Our method extracts disentangled representations of facial expressions, morphology, and poses via an expression decoupling module, and then maps the decoupled expression representations to the robot’s actuator command via an expression transfer module,  as is shown in Figure~\ref{pipeline_all}. Specifically,  the expression decoupling module employs an encoder–decoder architecture to learn disentangled latent representations in a fully self-supervised manner.  The encoder extracts separate expression and morphology representations from real face images, and the decoder, implemented using a statistical 3D head model FLAME~\cite{FLAME:SiggraphAsia2017}, reconstructs 3D faces conditioned on these representations.  Correspondences between the generated and real faces serve as self-supervision signals to guide the learning of disentangled representations without annotated data. 
We further introduce an expression transfer module to map the learned representations to the robot’s actuator commands through a learning objective of perceiving expression
errors. The module is conditioned on both the human expression and the robot’s intrinsic facial morphology, thereby reproducing faithful and human-like expression without requiring user-specific calibration..

We evaluate our method in both synthetic and real situations. 

A synthetic dataset is  constructed using the FLAME model, where each sample consists of a triplet: expression parameters, morphology parameters, and the corresponding rendered face image. 
Experimental results confirm that our method successfully disentangles facial expressions from morphology and consistently surpasses baseline methods. Furthermore, we develop Pengrui, a novel human-face robot tailored for high-fidelity expression imitation, to evaluate our method under real-world conditions. Pengrui pioneers the use of stepper motors for facial actuation, operating with $32$ actuators. %degrees of freedom (DoFs). 
These motors are directly connected to a silicone facial skin via rigid linkage structures. This unique hardware architecture endows the robot with significantly faster dynamic responses and a larger range of motion compared to traditional designs. Real-robot experiments demonstrate that, empowered by our morphology-independent method, Pengrui can accurately interpret and reproduce a wide array of nuanced human expressions with exceptional fidelity.

Our main contributions are as follows:
\begin{itemize}

\item We propose a novel morphology-independent facial expression imitation method for human-face robots. Our method explicitly decouples facial expressions from individual morphology in a self-supervised manner, which eliminates the interference of morphological variations, thereby enabling robots generate highly realistic and accurate robotic expressions.

\item We develop Pengrui, a human-face robot that serves as a dedicated experimental platform for our method. Pengrui pioneers previous human-face robots with its superior number of degrees of freedom (DoFs), faster response speed, and broader range of motion, which enables the validation of our facial expression imitation method. 

\end{itemize}

\section{RELATED WORKS}

\subsection{Facial Expression Imitation}

Facial expression imitation for human-face robots constitutes a key research area in human-robot interaction \cite{breazeal2003emotion,goodrich2008human,yan2014survey,dimitrievska2020behavior,song2009image,liu2017facial,gu2017local}, seeking to faithfully reproduce human facial expressions for natural and expressive human–robot interaction. Early approaches  often relied on pre-defined expression patterns to drive robotic imitation. For example, Wu et al.~\cite{wu2024retargeting} stored 12 canonical patterns of facial expressions on a physical robot and applied nearest-neighbour retrieval to select the closest pattern for actuation. More recent work shifted toward using sparse facial landmarks as a more generalizable representation for expressions. Several  studies~\cite{chen2021smile,ishihara2011realistic,hashimoto2008dynamic,zhang2025morpheus}  employed learning-based models to capture the motor behavior of human-face robots, producing general and flexible expression imitation. Eva 2.0 \cite{chen2021smile} and Emo \cite{hu2024human} utilized   MediaPipe \cite{lugaresi2019mediapipeframeworkbuildingperception} to extract 2D facial landmarks as expression  representation, while XIN-REN \cite{ren2016automatic} extended this idea by adopting 3D facial landmarks to mimic expressions from a specific individual.

Existing methods have achieved impressive imitation performance when facial morphology remains consistent. However, they are prone to expression distortion under morphological variations, as their expression representations are inherently coupled with facial morphology. In contrast, our morphology-independent expression imitation method explicitly decouples expressions from morphology, thereby mitigating morphological interference and enabling more robust facial reproduction.

\subsection{Human-Face Robot}

Human-face robots have a physical animatronic face with soft and human-like skin. Research in this area plays a crucial and irreplaceable role in promoting natural human–robot interaction~\cite{bartneck2024human}.  A variety of actuation strategies have been developed for constructing such robots.  For instance, Kobayashi et al.~\cite{kobayashi1993study} adopted flexible microactuators (FMAs) as the actuation mechanism, enabling robots to generate  diverse facial expressions. Li et al.~\cite{li2024driving} proposed a tendon-driven approach and applied linear blend skinning (LBS) for speech-synchronized facial animation. Recently, the rigid-body actuation strategy was used in several human-face robot designs~\cite{hu2024human}~\cite{chen2021smile}, and a hybrid actuation scheme that combines rigid-body and tendon-driven components was also introduced~\cite{ke2015facial} to enhance expressive capability.

While existing human-face robots demonstrate competent performance, most are proprietary in nature and not openly available. To evaluate our morphology-independent facial expression imitation method under real-world conditions, we  design and implement a novel human-face robot named Pengrui, based on a rigid-linkage actuation system. In contrast to prior platforms, Pengrui features a completely redesigned mechanical structure that supports enhanced expressivity and more accurate facial imitation.

\section{METHOD}

\subsection{Overview}

We propose a morphology-independent expression imitation method and validate it on Pengrui, a newly developed human-face robot, under real-world conditions.
Our method comprises two core components: an expression decoupling module and an expression transfer module. As illustrated in Fig.~\ref{pipeline_all}, given an input facial image  $\mathbf{I}\in \mathbb{R}^{W \times H \times 3}$ with a natural expression, the expression decoupling module extracts  disentangled representations of the target facial expression $\mathbf{e}$, head pose $\mathbf{p}$, and facial morphology $\mathbf{m}$, respectively. The representation $\mathbf{e}$ is then passed to the expression transfer module, generating the corresponding actuator control commands $\tilde{\mathbf{a}}$  for the human-face robot. These commands $\tilde{\mathbf{a}}$ are finally sent to  Pengrui to reproduce the intended facial expression, shown as output image $\mathbf{I}^r$ in Fig.~\ref{pipeline_all}.

\begin{figure*}[thpb]
    \centering
    \includegraphics[width=\textwidth]{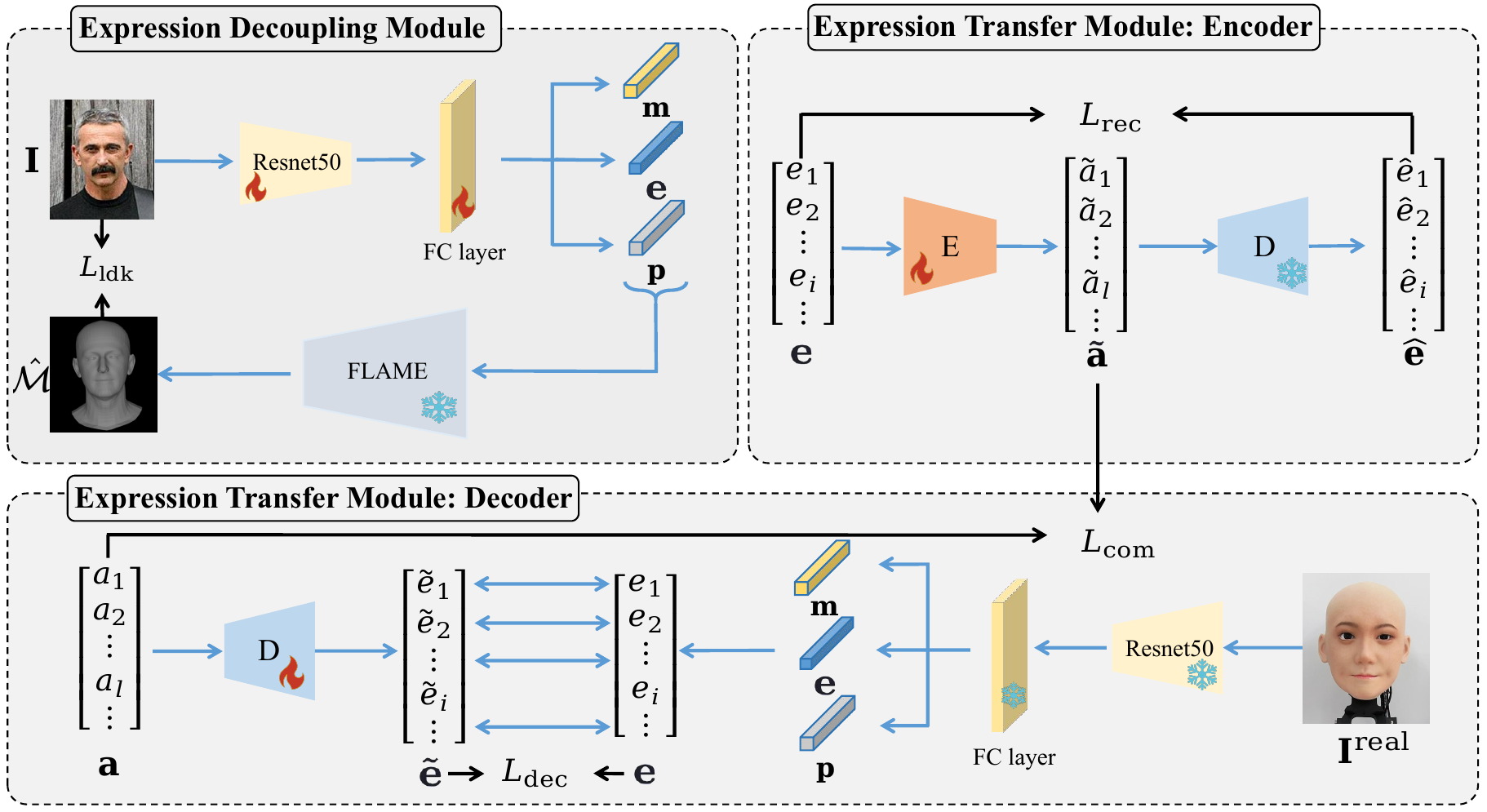}%\includegraphics[scale=1.0]{figurefile}
    \caption{\textbf{ Illustration of the expression decoupling module and the expression transfer module.} }
    \label{pipeline}
\end{figure*}

\subsection{Expression Decoupling Module} \label{expressiondecoupledmodel}

The expression decoupling module aims to decouple the expression of the imitated person from the facial morphology in the representation space.

Inspired by the ability of deep representation learning to implicitly separate semantic attributes, we  implement the module as a deep neural network, denoted as $\text{EDM}_\theta$ with parameters $\theta$. This module extracts the facial expression representation $\mathbf{e}= [e_1,e_2, \cdots,e_i,\cdots ]^\top$, head pose $\mathbf{p}= [p_1,p_2, \cdots, p_j, \cdots]^\top$, and facial morphology $\mathbf{m}=[m_1,m_2, \cdots, m_k,\cdots ]^\top$ by 
\begin{equation}
\mathbf{e},\mathbf{p},\mathbf{m}=\text{EDM}_\theta(\mathbf{I}).  \label{equ:resnet}
\end{equation}
As illustrated in Fig.~\ref{pipeline}, we employ a ResNet50\cite{he2015deepresiduallearningimage} backbone followed by a fully connected layer to construct $\text{EDM}_\theta$. 
 
 A key challenge in training $\text{EDM}_\theta$ is the absence of sufficient supervised signals. To this end, we introduce a self-supervised learning framework leveraging the FLAME~\cite{FLAME:SiggraphAsia2017} model, a statistical 3D head model.  FLAME  maps expression parameters $\mathbf{e}_f$, shape parameters $\mathbf{s}_f $ and pose parameters $\mathbf{p}_f$ to a linear shape space to generate a 3D face mesh ${\hat{\mathcal{M}} = (\textbf{V},\textbf{F})}$, where $\textbf{V}$ denotes  the vertex matrix and $\textbf{F}$ is the index matrix representing mesh connectivity.
 Crucially, we establish  direct correspondences between the FLAME’s parameters and the decoupled representations from $\text{EDM}_\theta$, i.e., $\mathbf{e}_f \leftrightarrow \mathbf{e}$, $\mathbf{p}_f \leftrightarrow \mathbf{p}$ and $\mathbf{s}_f \leftrightarrow \mathbf{m}$. 
 
We adopt an encoder-decoder architecture with the real images $\mathbf{I}$ as input and the generated 3D faces $\hat{\mathcal{M}}$ as output, thereby enabling self-supervised  training to an autoencoder.  Specifically,  we extract 3D landmarks $\hat{L}_{3D}=\{(\hat{x}_i,\hat{y}_i,\hat{z}_i)|i=0,1,\cdots,N\}$ from $\hat{\mathcal{M}}$, and project them onto 2D space to obtain 2D landmarks $\hat{L}_{2D}=\{(\hat{x}_i,\hat{y}_i)|i=0,1,\cdots,N\}$, where $N$ is the number of  landmarks.  Corresponding 2D landmarks $L_{2D}=\{(x_i,y_i)|i=0,1,\cdots,N\}$ are also detected from the input image $\mathbf{I}$. Since the FLAME-generated face is conditioned on the representations derived from $\mathbf{I}$, the input and output faces should exhibit consistent expression and morphology. We therefore minimize the difference  between  $L_{2D}$ and  $\hat{L}_{2D}$ using a landmark-based loss functio $ L_{\text {ldk}}$, 
 \begin{equation}
        L_{\text {ldk}} = \sum_{i=1}^{N} \left( 
\left| x_i - \left( s \, \hat{x}_i + t_x \right) \right| 
+ 
\left| y_i - \left( s \, \hat{y}_i + t_y \right) \right| 
\right),
\end{equation}
 where  $s$ denotes the scaling factor, and $t_x,t_y$ represent the translation components along the $x$ and $y$ axes, respectively.

\subsection{Expression Transfer Module} \label{etm}

The expression transformation module takes the decoupled expression representation $\mathbf{e}$ as input and generates  the robot's actuator commands $\tilde{\mathbf{a}}=[\tilde{a}_1,\tilde{a}_2,\cdots,\tilde{a}_l,\cdots]^\top$, as defined in 
\begin{equation}
    \tilde{\mathbf{a}}=\text{ETM}_\phi(\mathbf{e}), \label{equ:encoder}
\end{equation}
where $\phi$ is the model parameter. A total of $22$ actuator commands are used to control facial expressions on the human-face robot. Since no accurate physical model exists for this mapping, we implement $\text{ETM}_\phi$ using a fully-connected neural network to approximate the relationship between expression representations and actuator commands.

To train $\text{ETM}_\phi$, we collect a dataset comprising various expression images $\mathbf{I}$ and their corresponding representations $\mathbf{e}$ acquired in Eq. (\ref{equ:resnet}) by executing random actuator commands  $\mathbf{a}$ on the human-face robot. Each training sample is thus a pair $(\mathbf{a},\mathbf{e})$. Previous approaches such as \cite{hu2024human}  train the mapping $\text{ETM}_\phi$ by minimizing the error between the predicted command $\tilde{\mathbf{a}}$ in Eq. (\ref{equ:encoder}) and the  ground-truth command $\mathbf{a}$. However, minimizing command reconstruction error does not guarantee similarity in the resulting facial expressions. 

To directly optimize for perceptual expression fidelity, we introduce  an inverse expression transformation module, $\text{ETM-INV}_\psi$ with  parameters $\psi$, given by 
\begin{equation}
    \tilde{\mathbf{e}}=\text{ETM-INV}_\psi(\mathbf{a}),\label{equ:decoder}
\end{equation}
where  $\tilde{\mathbf{e}}$ is the predicted expression representation. $\text{ETM-INV}_\psi$ is also implemented as a fully-connected neural network. 
$\text{ETM}_\phi$ and $\text{ETM-INV}_\psi$ form an encoder-decoder architecture analogous to that used in the expression decoupling module. Specifically, the decoder  $\text{ETM-INV}_\psi$  predicts the expression representation $\tilde{\mathbf{e}}$ conditioned on the commands $\mathbf{a}$. The encoder $\text{ETM}_\phi$ maps the expression representation $\mathbf{e}$ to the robot's actuator commands $\mathbf{a}$. The entire system is trained by minimizing the expression reconstruction error between the original $\mathbf{e}$ and the reconstructed $\tilde{\mathbf{e}}$, thereby ensuring that the generated commands produce the intended facial expressions.

\noindent \textbf{Decoder.}
The decoder receives the robot's actuator commands $\mathbf{a}$ as input and predicts the robot's expression representation $\tilde{\mathbf{e}}$  in Eq. (\ref{equ:decoder}). To train the decoder, we collect a dataset of command–expression pairs from the physical robot.  We repeatedly sample real-world actuator commands $\mathbf{a}^{\text{real}}=[a_1^{\text{real}},a_2^{\text{real}},\cdots,a_l^{\text{real}},\cdots]^\top$ each of which produces a facial image $\mathbf{I}^{\text{real}}$ on the robot.  These commands and images are gathered over multiple trials and preprocessed before training. Specifically, we normalize each actuator’s input signal to the range $[0,1]$, due to that variations in the operating ranges of different actuators may result from their mechanical properties and spatial arrangement.
The normalization function is 
\begin{equation}
    \mathbf{a}  = \frac{\mathbf{a}^{\text{real}}-\min(\mathbf{a}^{\text{real}})}{\max(\mathbf{a}^{\text{real}})-\min(\mathbf{a}^{\text{real}})},
\end{equation}
where $\min(\cdot)$ and $\max(\cdot)$ indicate vectors composed of the minimum and maximum values of each actuator signal across all collected commands, respectively. The normalized command vector $\mathbf{a}$ is then fed to the decoder predict the expression representation via $\tilde{\mathbf{e}}=\text{ETM-INV}_\psi(\mathbf{a})$ in Eq.~(\ref{equ:decoder}).

To train the decoder, we obtain the ground-truth expression representation $\mathbf{e}$  from the collected facial images  $\mathbf{I}^{\text{real}}$.  As shown in Fig.~\ref{pipeline}, 
each $\mathbf{I}^{\text{real}}$  is passed through the pre-trained and frozen expression decoupling module $\text{EDM}_\theta$ from Eq. (\ref{equ:resnet}), yielding the corresponding expression representation $\mathbf{e}$ as the ground truth. This results in a training dataset consisting of pairs $(\mathbf{a},\mathbf{e})$ and the ground-truth $\mathbf{e}$, We define a decoder loss function $L_{dec}$ that measures the discrepancy between the predicted expression representation $\tilde{\mathbf{e}}$ in Eq. (\ref{equ:decoder}). The network parameters are optimized using the Adam algorithm by minimizing
\begin{equation}
    L_{dec}  =  \|\mathbf{e}-\tilde{\mathbf{e}} \|^2. 
\end{equation}

\noindent \textbf{Encoder.} The encoder $\text{ETM}_\phi$ is designed symmetrically to the decoder. It takes the expression representation $\mathbf{e}$ as input and predicts the corresponding actuator commands $\tilde{\mathbf{a}}$ in Eq. (\ref{equ:encoder}). Using the same sampled command–expression pairs  $(\mathbf{a},\mathbf{e})$, we compute an L2 loss $L_{\text{com}}$ between the predicted command $\tilde{\mathbf{a}}$ and the ground truth $\mathbf{a}$. $L_{\text{com}}$ is 
\begin{equation}
    L_{\text{com}}  =  \|\mathbf{a}-\tilde{\mathbf{a}} \|^2. 
\end{equation}
As previously noted, minimizing command error does not ensure perceptual similarity in the resulting expressions. To better align the robot’s expressions with human-like outputs, we further propagate the encoder’s output through the frozen and  pre-trained decoder to reconstruct the expression representation 
$\hat{\mathbf{e}}$ via  
\begin{equation}
    \hat{\mathbf{e}}=\text{ETM-INV}_\psi(\text{ETM}_\phi(\mathbf{e})).\label{equ:autoencoder}
\end{equation}
An expression reconstruction loss $L_{\text{rec}}$ is then computed as the L2 distance between $\hat{\mathbf{e}}$ and the ground truth  $\mathbf{e}$, given by 
\begin{equation}
    L_{\text{rec}}  =  \|\mathbf{e}-\hat{\mathbf{e}} \|^2. 
\end{equation}
The total loss function for training the encoder
\begin{equation}
    L_{\text{enc}}  =   L_{\text{com}} +\lambda L_{\text{rec}},
\end{equation}
where $\lambda = 1$ is the trade-off parameter. The encoder is optimized using the Adam algorithm under this composite objective, while the decoder remains fixed during training.

\section{EXPERIMENTS}

% \subsection{Implementation Details}
\subsection{Datasets and Implementation Details}
% \textbf{Datasets.} 
To train the expression decoupling module, we use the publicly available VGGFace2 dataset\cite{cao2018vggface2datasetrecognisingfaces}. VGGFace2 contains images of over $8k$ subjects, each with an average of over  $350$ images. We annotate the expression labels for VGGFace2 using the MAE-DFER facial expression recognition model~\cite{sun2023mae} for evaluation purposes.  Additionally, we construct a synthetic facial dataset using the FLAME model, where each sample is a tuple  $((\mathbf{e},\mathbf{p},\mathbf{m}),\mathcal{M})$, and we generate $1,000$ samples in total. To train the expression transfer module, as described in Section~\ref{etm}, we collect a robot-specific dataset consisting of command–image   pairs $(\mathbf{a},\mathbf{I}^{\text{real}})$.  The dataset is split into training and testing sets with an  $80\%/20\%$ ratio. % 看附录

The expression decoupling module $\text{EDM}_\theta$ was trained for $500$ epochs with a learning rate of $0.0001$ and a batch size of $16$. For the expression transfer module  $\text{ETM}_\phi$, both the encoder and decoder were optimized with a learning rate of $0.001$ and a batch size of $32$. All models were trained using the Adam optimizer. All experiments were performed on a single NVIDIA RTX 4090 GPU.

\begin{figure}[t]
    \centering
    \includegraphics[width=\linewidth]{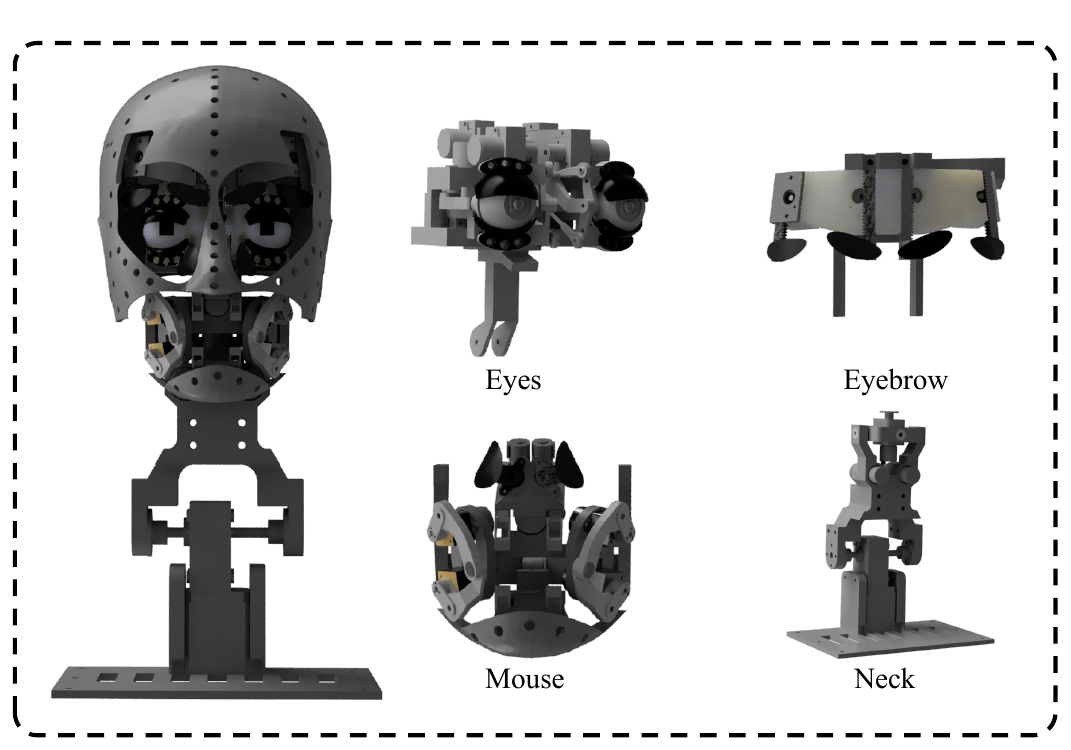} %\includegraphics[scale=1.0]{figurefile}
    \caption{\textbf{Pengrui, a highly articulated human-face robot platform.} Pengrui robot has $32$ actuators with $48$ degrees of freedom.}
    \label{figurepengrui}
\end{figure}
\subsection{Robot Design}
We developed Pengrui, a novel human-face robotic platform dedicated to evaluating facial expression imitation methods, as is shown in Fig.~\ref{figurepengrui}.  The platform employs a rigid-linkage actuation mechanism, utilizing stepper motors to manipulate subdermal anchors embedded within a silicone skin.

A total of $23$ actuators are dedicated to facial expressions, covering key regions (brows, nasolabial, lips, chin, jaw, cheeks), while $9$ actuators control eye and neck movements for integrated gaze and head gestures. Fig.~\ref{figurepengrui} illustrates the actuator distribution, which balances facial coverage with controllability and learning identifiability.

\noindent\textbf{Eye Module.} A compact binocular assembly mounts two spherical eyeballs on a shared pan–tilt gimbal for yoked gaze; a differential linkage introduces vergence without disrupting global alignment. Grouped eyelid actuation yields controllable aperture and blink dynamics. This module enhances realism and supports coordination cues (gaze shift with smile or surprise).

\noindent\textbf{Mouth Module.} The mouth mechanism combines corner pullers/depressors, midline upper/lower actuators, and bilateral stretchers/puckers. Short, low–friction linkages couple actuators to subdermal pads around the vermilion border, enabling independent control of vertical (smile/frown), horizontal (stretch), and protrusion (pucker) components. The jaw actuator modulates aperture and supports coarticulation with lip shaping.

\noindent\textbf{Neck Module.} A four-DoF neck mechanism (incorporating pitch, yaw, and roll) enables a comprehensive range of expressive head movements—such as nodding, turning, and lateral tilting—that temporally synchronize with facial changes. Mechanical limits and compliant couplers protect the silicone–skull interface and ensure safe operation near range extremes.

\noindent\textbf{Materials, Sensing, and Calibration.} The face shell is poured from a medium–soft silicone with graded thickness. Anchor pads are embedded during molding and keyed to the internal skull to prevent drift. Each actuator exposes high–resolution position feedback; an IMU in the head facilitates motion compensation during camera–based observation. Calibration proceeds by: (i) homing to the neutral pose, (ii) sweeping each actuator to identify $(a^{i}{\min}, a^{i}{\max})$, and (iii) recording camera–observed facial landmarks and mesh reprojection to align anchor influence with the FLAME expression space. These ranges define the per–actuator normalization used by the expression transfer module.

\subsection{Evaluation Metrics}
We employ Mean Squared Error (MSE), Mean Absolute Error (MAE), and the Coefficient of Variation (CV) to quantitatively evaluate our method. MSE and MAE are used to assess the prediction accuracy of both the expression decoupling and expression transfer modules, where lower values indicate higher predictive precision. The CV metric is introduced to compare the variability between landmark-based representations and our proposed morphology-independent representations across identical expressions. A lower CV reflects greater stability and robustness of the representation against variations in facial morphology.

\subsection{Baselines}

For the expression decoupling module, we use landmark-based representations as a baseline to evaluate the stability of our morphology-independent representations across varying facial morphologies under consistent expressions. Additionally, inspired by \cite{hu2024human}, we employ a random generation (RG) baseline that produces synthetic expression and morphology representations for error comparison with those predicted by our method in Eq.~(\ref{equ:resnet}).

For the expression transfer module, following~\cite{chen2021smile}, we compare against three baseline methods for predicting actuator commands, with error measured against our method (Eq.~(\ref{equ:encoder})). The first baseline uses a randomly-initialized version of our network (RI). The second employs the same network fine-tuned over $10$ optimization cycles (RI-10). The third adopts a Nearest Neighbor (NN) strategy~\cite{cover1967nearest}, which retrieves the command from the most similar training sample as the prediction.

\begin{table*}[htb]
    \centering
    \small
    \caption{CV comparisons of Representations across Different Face Morphologies under the Same Expression}
    \label{tab:landmarkandexpression}
    \begin{tabular}{l >{\centering\arraybackslash}p{1cm}>{\centering\arraybackslash}p{1cm}>{\centering\arraybackslash}p{1cm}>{\centering\arraybackslash}p{1cm}>{\centering\arraybackslash}p{1cm}>{\centering\arraybackslash}p{1cm}>{\centering\arraybackslash}p{1cm}}
        \toprule
        Representation & surprise$\downarrow$ & fear$\downarrow$ & disgust$\downarrow$ & happy$\downarrow$ & sad$\downarrow$ & anger$\downarrow$ & neutral$\downarrow$\\
        \midrule
        landmark-based         & 8.128 & 8.490 & 10.228 & 7.480 & 8.469 & 8.202 & 8.031 \\
         morphology-independent (ours) & \textbf{4.442} & \textbf{6.857} &\textbf{4.636} & \textbf{4.874}& \textbf{8.282}& \textbf{5.471} & \textbf{7.849} \\
        \bottomrule
    \end{tabular}

\end{table*}

\begin{table*}[ht]
    \centering
    \small
    \caption{MSE and MAE comparisons of Representations Errors}
    \label{tab:exp_morph_distance}
    \begin{tabular}{l >{\centering\arraybackslash}p{1.8cm}>{\centering\arraybackslash}p{1.8cm}>{\centering\arraybackslash}p{1.8cm}>{\centering\arraybackslash}p{1.8cm}>{\centering\arraybackslash}p{1.8cm}>{\centering\arraybackslash}p{1.8cm}}
        \toprule
        \multirow{2}{*}{Method} 
        & \multicolumn{2}{c}{Overall Representation}
        & \multicolumn{2}{c}{Expression Representation}
        & \multicolumn{2}{c}{Morphology Representation} \\
        \cmidrule(lr){2-3} \cmidrule(lr){4-5} \cmidrule(lr){6-7}
        & MSE $\downarrow$ & MAE $\downarrow$ & MSE $\downarrow$ & MAE $\downarrow$ & MSE $\downarrow$ & MAE $\downarrow$ \\
        \midrule
        RG         & 3.838 & 1.612 & 4.001 & 1.669 & 3.981 & 1.663 \\

        EDM (ours) & \textbf{1.006} & \textbf{0.794} & \textbf{1.108} & \textbf{0.840} & \textbf{1.012} & \textbf{0.806} \\
        \bottomrule
    \end{tabular}
\end{table*}

\begin{table}[ht]
    \centering
    \small
    \caption{MSE and MAE comparisons of Command Errors}
    \label{tab:eum_variants}
    \begin{tabular}{lcc}
        \toprule
        Method & MSE $\downarrow$ & MAE $\downarrow$ \\
        \midrule
        EDM + RI     & 0.314 & 0.479 \\
        EDM + RI-10 & 0.089 & 0.283 \\
        EDM + NN    & 0.101 & 0.261 \\
        RG + ETM    & 0.303 & 0.465 \\
        EDM + ETM (ours)   & \textbf{0.042} & \textbf{0.174} \\
        \bottomrule
    \end{tabular}
\end{table}

\subsection{Results}
\noindent \textbf{Expression Decoupling Module.} Tab.~\ref{tab:landmarkandexpression} 
% ,Tab.~\ref{tab:rep_distance} 
and Tab.~\ref{tab:exp_morph_distance} show the quantitative evaluations for our expression decoupling module.

We categorize facial images by expression and compute the coefficient of variation (CV) for both landmark-based representations and our morphology-independent representations within each category. The results are presented in Tab.~\ref{tab:landmarkandexpression}. Our method achieves the most notable improvement for surprise and disgust expressions, reducing CV by $3.686$ and $5.592$, respectively. In contrast, improvements for sad and neutral expressions are relatively modest. We attribute this to the inherently high intra-class variability of sad and neutral expressions in natural human behavior, which are often more subtle, person-specific, and less distinctive than prototypical expressions such as happiness or surprise. 
This observation underscores a promising direction for future work: enhancing the sensitivity of expression representations to subtle emotional cues and further reducing intra-class variation for less visually distinct categories.

\begin{figure}[!t]
    \centering
    \includegraphics[width=\linewidth]{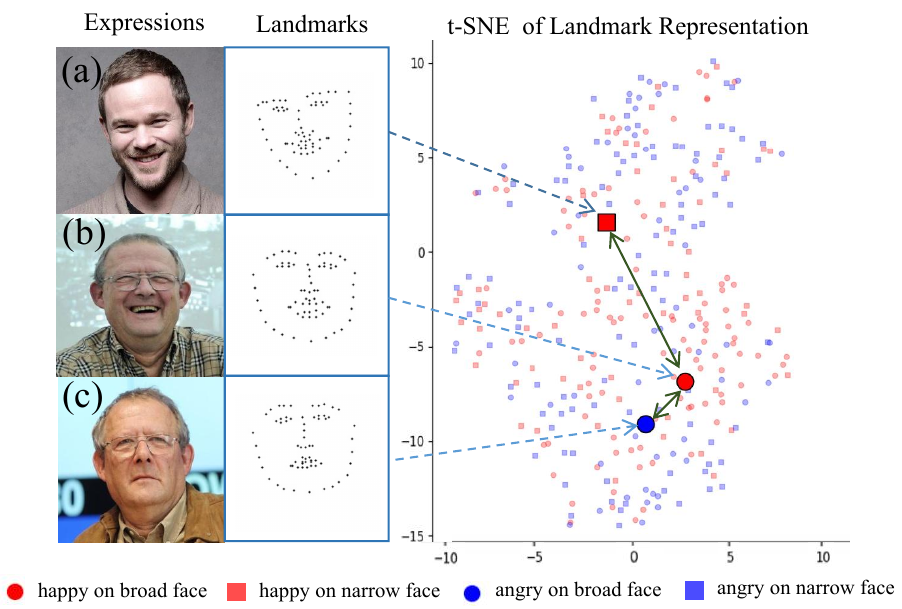 } %\includegraphics[scale=1.0]{figurefile}
    \caption{\textbf{Examples of interferences from facial morphology.} The 2D facial landmarks for the (a) happy expression on a narrow face, (b) happy expression on a broad face and (c) angry expression on a broad face, are mapped to low-dimensional space using the t-SNE algorithm. The distance for the same expression across different morphology shown in (a) and (b) significantly exceeds that for different expressions on the same morphology shown in (a) and (c).  }
    \label{figurelabel}
\end{figure}

Fig.~\ref{figurelabel} illustrates how landmark-based representations inherently entangle expression and facial morphology. Specifically, the landmark variance for the same expression across different morphologies (e.g., happy in Fig.~\ref{figurelabel}(a) and Fig.~\ref{figurelabel}(b)) can exceed the variance between distinct expressions on the same face (e.g., happy and angry in Fig.~\ref{figurelabel}(a) and Fig.~\ref{figurelabel}(c)). Consequently, slight morphological changes easily disrupt the mapping process, causing existing methods to misinterpret expressively distinct inputs and generate distorted robotic actuation.

\begin{figure}[h]
    \centering
    \includegraphics[width=\linewidth]{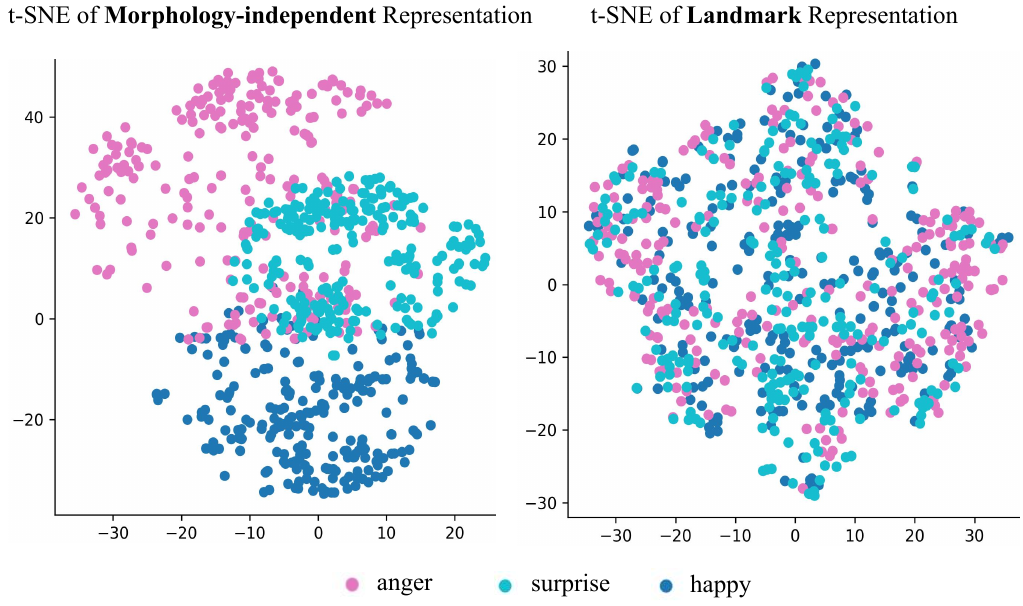} %\includegraphics[scale=1.0]{figurefile}
    \caption{\textbf{Representation Visualizations.} t-SNE visualization comparing the morphology-independent representation (our method) and landmark-based representation.}
    \label{exp_tsne}
\end{figure}

In contrast, our morphology-independent approach effectively overcomes this confounding effect. As visualized in the t-SNE [40] embeddings (Fig.~\ref{exp_tsne}), our decoupled representations form significantly tighter, distinct clusters for each expression across diverse morphologies. Meanwhile, landmark-based features exhibit severe scattering. These comparative results clearly confirm that our method successfully eliminates morphological bias, ensuring highly robust expression imitation.

\begin{figure*}[htb]
    \centering
    \includegraphics[width=\textwidth]{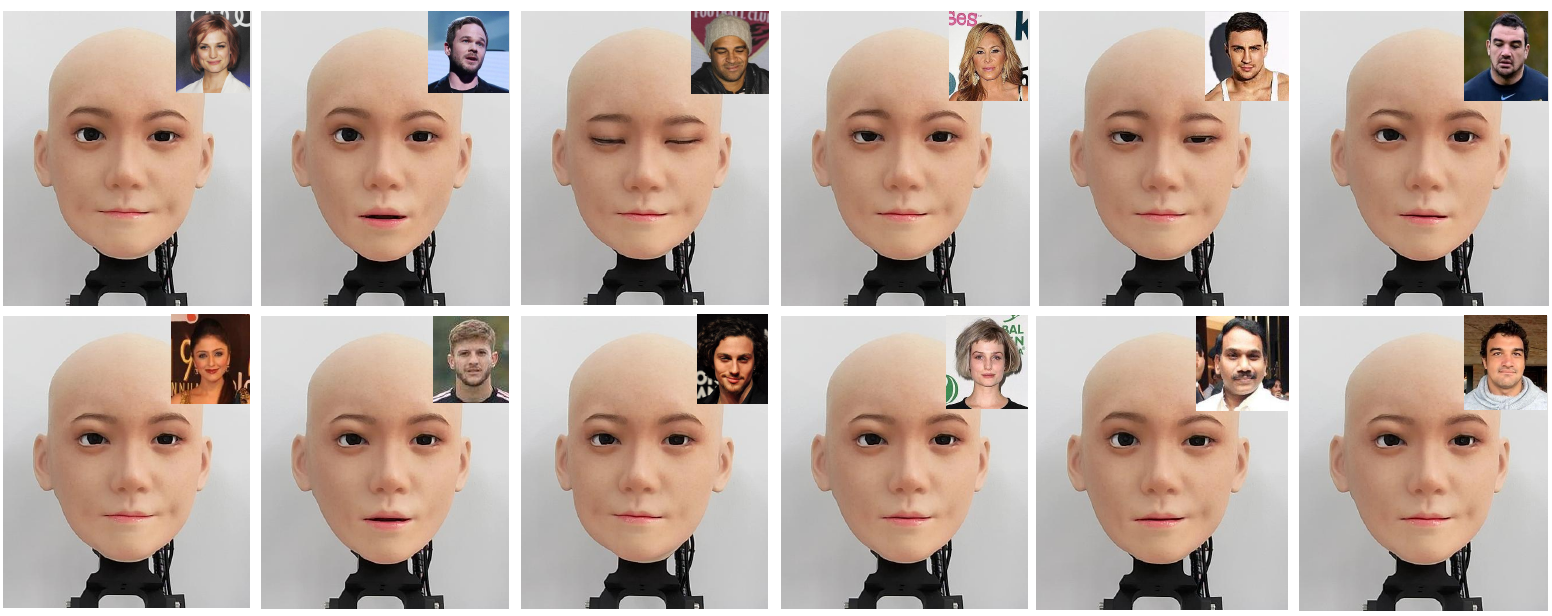}%\includegraphics[scale=1.0]{figurefile}
    \caption{\textbf{Robot Visualizations.}     The actuator commands $\tilde{\mathbf{a}}$ generated by our method are deployed on the human-face robot. The robot accurately reproduces facial expressions from different individuals, demonstrating the effectiveness and generalizability of our approach. }
    \label{expriment_robot}
\end{figure*}
We evaluate the performance of our method by computing the Mean Squared Error (MSE) and Mean Absolute Error (MAE) between the decoupled representations $(\mathbf{e},\mathbf{p},\mathbf{m})$ and the  ground truth. Quantitative results are summarized in Tab.~\ref{tab:exp_morph_distance}. Our method consistently outperforms the baseline across all metrics. Specifically, for the overall representation, our approach reduces MSE by $2.832$ ($73.8\%$) and MAE by $0.818$ ($50.7\%$). For expression representation alone, it achieves an MSE reduction of $2.893$ ($72.3\%$) and an MAE reduction of $0.829$ ($49.7\%$). In morphology representation, the method attains an MSE reduction of $2.969$ ($74.6\%$) and an MAE reduction of $0.857$ ($51.5\%$). These results confirm the effectiveness of our method in learning accurate and disentangled expression and morphology representations.

\noindent \textbf{Expression Transfer Module and Entire Framework.} We integrate our expression decoupling module with multiple baseline variants of the expression transfer module and test the full pipeline on our robot-specific dataset. To quantify imitation accuracy, we measure the MSE and MAE between the predicted actuator commands and the ground-truth values.

As summarized in Tab.~\ref{tab:eum_variants}, our method (EDM+ETM) consistently outperforms all baseline combinations including EDM+RI, EDM+RI-10, and EDM+NN. Compared to the strongest baseline (EDM+NN), our approach reduces MSE by $0.059$ ($58.4\%$) and MAE by $0.073$ ($33.3\%$). These results confirm the effectiveness of our expression transfer module in accurately generating actuator commands and underscore the complementary role of both modules within the overall pipeline.

To evaluate the importance of the expression decoupling module (EDM), we replace it with the random generation (RG) baseline while retaining our expression transfer module (ETM), and compute the MSE and MAE between predicted and ground-truth actuator commands. As shown in Tab.~\ref{tab:eum_variants}, using RG+ETM leads to a significant performance drop: MSE increases by $0.261$ (from $0.042$ to $0.303$) and MAE by $0.291$ (from $0.174$ to $0.465$). These results confirm the essential role of our expression decoupling module in achieving high-fidelity expression imitation.

\noindent \textbf{Real-World Execution. } 
We deploy our method on the physical human-face robot platform to evaluate its real-world performance. Qualitative results, illustrated in Fig.~\ref{expriment_robot}, show that our approach enables the robot to accurately capture expressions from individuals with diverse facial morphology and reproduce them consistently on the own face. These outcomes demonstrate the stability and generalizability of our method in practical human–robot interaction scenarios.

\subsection{Limitations}
Despite the demonstrated effectiveness, our method shows relatively lower fidelity in subtle expressions such as surprise and disgust. This limitation arises from the difficulty of capturing fine-grained emotional cues within morphology-independent representations. Furthermore, practical factors such as the long-term stability of silicone materials and the durability of actuators may lead to expression drift, thereby reducing consistency and undermining the robustness of self-supervised training over time.

\section{CONCLUSION AND FUTURE WORK}

We have presented a morphology-independent method that can decouple expressions from facial morphology for high-fidelity facial-expression imitation for human-face robots. Our method  consists of two core components: an expression decoupling module that can learn to disentangle expression and morphology representations in a self-supervised manner, and an expression transfer module that can map these representations to actuator commands by minimizing perceptual expression errors.  The two modules coordinate with each other, which can finally eliminate the influence of facial morphology and produce more realistic expressions on human-face robots. To support real-world validation, we developed a novel and expressive human-face robot as an experimental platform. Extensive experimental results demonstrate the effectiveness and generalizability of our method.

In the future, we plan to enhance representation learning to better capture subtle emotional cues and establish improved evaluation protocols, including human perceptual studies. In parallel, upgrades to Pengrui's materials, actuators and control strategies will aim to improve stability, repeatability, and naturalness in imitation of expression, allowing more expressive and reliable human–robot interaction.

\bibliographystyle{IEEEtran}
\bibliography{IEEEtranBST/IEEEabrv,IEEEtranBST/reference}

@article{pennisi2016autism,
  title={Autism and social robotics: A systematic review},
  author={Pennisi, Paola and Tonacci, Alessandro and Tartarisco, Gennaro and Billeci, Lucia and Ruta, Liliana and Gangemi, Sebastiano and Pioggia, Giovanni},
  journal={Autism Research},
  volume={9},
  number={2},
  pages={165--183},
  year={2016},
  publisher={Wiley Online Library}
}

@article{fogelson2022impact,
  title={The impact of robotic companion pets on depression and loneliness for older adults with dementia during the COVID-19 pandemic},
  author={Fogelson, Donna M and Rutledge, Carolyn and Zimbro, Kathie S},
  journal={Journal of Holistic Nursing},
  volume={40},
  number={4},
  pages={397--409},
  year={2022},
  publisher={SAGE Publications Sage CA: Los Angeles, CA}
}

@inproceedings{10.1145/3568162.3578625,
author = {Jeong, Sooyeon and Aymerich-Franch, Laura and Alghowinem, Sharifa and Picard, Rosalind W. and Breazeal, Cynthia L. and Park, Hae Won},
title = {A Robotic Companion for Psychological Well-being: A Long-term Investigation of Companionship and Therapeutic Alliance},
year = {2023},
isbn = {9781450399647},
publisher = {Association for Computing Machinery},
doi = {10.1145/3568162.3578625},
booktitle = {Proceedings of the 2023 ACM/IEEE International Conference on Human-Robot Interaction},
pages = {485–494},
numpages = {10},
series = {HRI '23}
}

@article{laban2025coping,
  title={Coping with emotional distress via self-disclosure to robots: An intervention with caregivers},
  author={Laban, Guy and Morrison, Val and Kappas, Arvid and S. Cross, Emily},
  journal={International Journal of Social Robotics},
  pages={1--34},
  year={2025},
  publisher={Springer}
}

@article{FLAME:SiggraphAsia2017, 
  title = {Learning a model of facial shape and expression from {4D} scans}, 
  author = {Li, Tianye and Bolkart, Timo and Black, Michael. J. and Li, Hao and Romero, Javier}, 
  journal = {ACM Transactions on Graphics, (Proc. SIGGRAPH Asia)}, 
  volume = {36}, 
  number = {6}, 
  year = {2017}, 
  pages = {194:1--194:17},

}

@inproceedings{oh2006design,
  title={Design of android type humanoid robot Albert HUBO},
  author={Oh, Jun-Ho and Hanson, David and Kim, Won-Sup and Han, Young and Kim, Jung-Yup and Park, Ill-Woo},
  booktitle={2006 IEEE/RSJ International Conference on Intelligent Robots and Systems},
  pages={1428--1433},
  year={2006},
  organization={IEEE}
}

@article{hu2024human,
  title={Human-robot facial coexpression},
  author={Hu, Yuhang and Chen, Boyuan and Lin, Jiong and Wang, Yunzhe and Wang, Yingke and Mehlman, Cameron and Lipson, Hod},
  journal={Science Robotics},
  volume={9},
  number={88},
  pages={eadi4724},
  year={2024},
  publisher={American Association for the Advancement of Science}
}

@inproceedings{li2024driving,
  title={Driving Animatronic Robot Facial Expression From Speech},
  author={Li, Boren and Li, Hang and Liu, Hangxin},
  booktitle={2024 IEEE/RSJ International Conference on Intelligent Robots and Systems (IROS)},
  pages={7012--7019},
  year={2024},
  organization={IEEE}
}

@inproceedings{chen2021smile,
  title={Smile like you mean it: Driving animatronic robotic face with learned models},
  author={Chen, Boyuan and Hu, Yuhang and Li, Lianfeng and Cummings, Sara and Lipson, Hod},
  booktitle={2021 IEEE International Conference on Robotics and Automation (ICRA)},
  pages={2739--2746},
  year={2021},
  organization={IEEE}
}

@article{ren2016automatic,
  title={Automatic facial expression learning method based on humanoid robot XIN-REN},
  author={Ren, Fuji and Huang, Zhong},
  journal={IEEE Transactions on Human-Machine Systems},
  volume={46},
  number={6},
  pages={810--821},
  year={2016},
  publisher={IEEE}
}

@article{zhang2025morpheus,
  title={Morpheus: A Neural-driven Animatronic Face with Hybrid Actuation and Diverse Emotion Control},
  author={Zhang, Zongzheng and Yang, Jiawen and Peng, Ziqiao and Yang, Meng and Ma, Jianzhu and Cheng, Lin and Xu, Huazhe and Zhao, Hang and Zhao, Hao},
  journal={arXiv preprint arXiv:2507.16645},
  year={2025}
}

@book{bartneck2024human,
  title={Human-robot interaction: An introduction},
  author={Bartneck, Christoph and Belpaeme, Tony and Eyssel, Friederike and Kanda, Takayuki and Keijsers, Merel and {\v{S}}abanovi{\'c}, Selma},
  year={2024},
  publisher={Cambridge University Press}
}

@inproceedings{kobayashi1993study,
  title={Study on face robot for active human interface-mechanisms of face robot and expression of 6 basic facial expressions},
  author={Kobayashi, Hiroshi and Hara, Fumio},
  booktitle={Proceedings of 1993 2nd IEEE International Workshop on Robot and Human Communication},
  pages={276--281},
  year={1993},
  organization={IEEE}
}

@inproceedings{ke2015facial,
  title={Facial expression on robot SHFR-III based on head-neck coordination},
  author={Ke, Xianxin and Yang, Yang and Xin, Jizhong},
  booktitle={2015 IEEE International Conference on Information and Automation},
  pages={1622--1627},
  year={2015},
  organization={IEEE}
}

@article{he2015deepresiduallearningimage,
  title={Deep Residual Learning for Image Recognition},
  author={He, Kaiming and Zhang, Xiangyu and Ren, Shaoqing and Sun, Jian},
  journal={arXiv preprint arXiv:1512.03385},
  year={2015}
}

@article{cao2018vggface2datasetrecognisingfaces,
  title={VGGFace2: A dataset for recognising faces across pose and age},
  author={Cao, Qiong and Shen, Li and Xie, Weidi and Parkhi, Omkar M. and Zisserman, Andrew},
  journal={arXiv preprint arXiv:1710.08092},
  year={2018}
}

@inproceedings{hashimoto2006development,
  title={Development of the face robot SAYA for rich facial expressions},
  author={Hashimoto, Takuya and Hitramatsu, Sachio and Tsuji, Toshiaki and Kobayashi, Hiroshi},
  booktitle={2006 SICE-ICASE International Joint Conference},
  pages={5423--5428},
  year={2006},
  organization={IEEE}
}

@article{liang2025largemodelempoweredembodied,
  title={Large Model Empowered Embodied AI: A Survey on Decision-Making and Embodied Learning},
  author={Liang, Wenlong and Zhou, Rui and Ma, Yang and Zhang, Bing and Li, Songlin and Liao, Yijia and Kuang, Ping},
  journal={arXiv preprint arXiv:2508.10399},
  year={2025}
}

@article{hess2013emotional,
  title={Emotional mimicry as social regulation},
  author={Hess, Ursula and Fischer, Agneta},
  journal={Personality and social psychology review},
  volume={17},
  number={2},
  pages={142--157},
  year={2013},
  publisher={Sage Publications Sage CA: Los Angeles, CA}
}

@article{iacoboni2005neural,
  title={Neural mechanisms of imitation},
  author={Iacoboni, Marco},
  journal={Current opinion in neurobiology},
  volume={15},
  number={6},
  pages={632--637},
  year={2005},
  publisher={Elsevier}
}

@article{breazeal2003emotion,
  title={Emotion and sociable humanoid robots},
  author={Breazeal, Cynthia},
  journal={International journal of human-computer studies},
  volume={59},
  number={1-2},
  pages={119--155},
  year={2003},
  publisher={Elsevier}
}

@article{goodrich2008human,
  title={Human--robot interaction: a survey},
  author={Goodrich, Michael A and Schultz, Alan C and others},
  journal={Foundations and trends{\textregistered} in human--computer interaction},
  volume={1},
  number={3},
  pages={203--275},
  year={2008},
  publisher={Now Publishers, Inc.}
}

@article{yan2014survey,
  title={A survey on perception methods for human--robot interaction in social robots},
  author={Yan, Haibin and Ang Jr, Marcelo H and Poo, Aun Neow},
  journal={International Journal of Social Robotics},
  volume={6},
  number={1},
  pages={85--119},
  year={2014},
  publisher={Springer}
}

@article{dimitrievska2020behavior,
  title={Behavior models of emotion-featured robots: A survey},
  author={Dimitrievska, Vesna and Ackovska, Nevena},
  journal={Journal of Intelligent \& Robotic Systems},
  volume={100},
  number={3},
  pages={1031--1053},
  year={2020},
  publisher={Springer}
}

@article{song2009image,
  title={Image ratio features for facial expression recognition application},
  author={Song, Mingli and Tao, Dacheng and Liu, Zicheng and Li, Xuelong and Zhou, Mengchu},
  journal={IEEE Transactions on Systems, Man, and Cybernetics, Part B (Cybernetics)},
  volume={40},
  number={3},
  pages={779--788},
  year={2009},
  publisher={IEEE}
}

@article{liu2017facial,
  title={A facial expression emotion recognition based human-robot interaction system.},
  author={Liu, Zhentao and Wu, Min and Cao, Weihua and Chen, Luefeng and Xu, Jianping and Zhang, Ri and Zhou, Mengtian and Mao, Junwei},
  journal={IEEE CAA J. Autom. Sinica},
  volume={4},
  number={4},
  pages={668--676},
  year={2017}
}

@article{gu2017local,
  title={Local robust sparse representation for face recognition with single sample per person},
  author={Gu, Jianquan and Hu, Haifeng and Li, Haoxi},
  journal={IEEE/CAA Journal of Automatica Sinica},
  volume={5},
  number={2},
  pages={547--554},
  year={2017},
  publisher={IEEE}
}

@inproceedings{wu2024retargeting,
  title={Retargeting human facial expression to human-like robotic face through neural network surrogate-based optimization},
  author={Wu, Bowen and Liu, Chaoran and Ishi, Carlos T and Minato, Takashi and Ishiguro, Hiroshi},
  booktitle={2024 IEEE/RSJ International Conference on Intelligent Robots and Systems (IROS)},
  pages={4724--4730},
  year={2024},
  organization={IEEE}
}

@inproceedings{ishihara2011realistic,
  title={Realistic child robot “affetto” for understanding the caregiver-child attachment relationship that guides the child development},
  author={Ishihara, Hisashi and Yoshikawa, Yuichiro and Asada, Minoru},
  booktitle={2011 ieee international conference on development and learning (icdl)},
  volume={2},
  pages={1--5},
  year={2011},
  organization={IEEE}
}

@inproceedings{hashimoto2008dynamic,
  title={Dynamic display of facial expressions on the face robot made by using a life mask},
  author={Hashimoto, Takuya and Hiramatsu, Sachio and Kobayashi, Hiroshi},
  booktitle={Humanoids 2008-8th IEEE-RAS International Conference on Humanoid Robots},
  pages={521--526},
  year={2008},
  organization={IEEE}
}

@misc{lugaresi2019mediapipeframeworkbuildingperception,
      title={MediaPipe: A Framework for Building Perception Pipelines}, 
      author={Camillo Lugaresi and Jiuqiang Tang and Hadon Nash and Chris McClanahan and Esha Uboweja and Michael Hays and Fan Zhang and Chuo-Ling Chang and Ming Guang Yong and Juhyun Lee and Wan-Teh Chang and Wei Hua and Manfred Georg and Matthias Grundmann},
      year={2019},
      eprint={1906.08172},
      archivePrefix={arXiv},
      primaryClass={cs.DC}
}

@article{wang2024disentangled,
  title={Disentangled representation learning},
  author={Wang, Xin and Chen, Hong and Tang, Si'ao and Wu, Zihao and Zhu, Wenwu},
  journal={IEEE Transactions on Pattern Analysis and Machine Intelligence},
  volume={46},
  number={12},
  pages={9677--9696},
  year={2024},
  publisher={IEEE}
}

@article{carbonneau2022measuring,
  title={Measuring disentanglement: A review of metrics},
  author={Carbonneau, Marc-Andr{\'e} and Zaidi, Julian and Boilard, Jonathan and Gagnon, Ghyslain},
  journal={IEEE transactions on neural networks and learning systems},
  volume={35},
  number={7},
  pages={8747--8761},
  year={2022},
  publisher={IEEE}
}

@article{hsieh2018learning,
  title={Learning to decompose and disentangle representations for video prediction},
  author={Hsieh, Jun-Ting and Liu, Bingbin and Huang, De-An and Fei-Fei, Li F and Niebles, Juan Carlos},
  journal={Advances in neural information processing systems},
  volume={31},
  year={2018}
}

@article{xu2022compositional,
  title={Compositional generalization in unsupervised compositional representation learning: A study on disentanglement and emergent language},
  author={Xu, Zhenlin and Niethammer, Marc and Raffel, Colin A},
  journal={Advances in Neural Information Processing Systems},
  volume={35},
  pages={25074--25087},
  year={2022}
}

@article{xie2024graph,
  title={Graph-based unsupervised disentangled representation learning via multimodal large language models},
  author={Xie, Baao and Chen, Qiuyu and Wang, Yunnan and Zhang, Zequn and Jin, Xin and Zeng, Wenjun},
  journal={Advances in Neural Information Processing Systems},
  volume={37},
  pages={103101--103130},
  year={2024}
}

@inproceedings{abbasi2024deciphering,
  title={Deciphering the role of representation disentanglement: Investigating compositional generalization in clip models},
  author={Abbasi, Reza and Rohban, Mohammad Hossein and Baghshah, Mahdieh Soleymani},
  booktitle={European Conference on Computer Vision},
  pages={35--50},
  year={2024},
  organization={Springer}
}

@article{sun2023mae,
  title={MAE-DFER: Efficient Masked Autoencoder for Self-supervised Dynamic Facial Expression Recognition},
  author={Sun, Licai and Lian, Zheng and Liu, Bin and Tao, Jianhua},
  journal={arXiv preprint arXiv:2307.02227},
  year={2023}
}

@article{cover1967nearest,
  title={Nearest neighbor pattern classification},
  author={Cover, Thomas and Hart, Peter},
  journal={IEEE transactions on information theory},
  volume={13},
  number={1},
  pages={21--27},
  year={1967},
  publisher={IEEE}
}
\end{document}